# Intelligent Singularity Avoidance in UR10 Robotic Arm Path Planning Using Hybrid Fuzzy Logic and Reinforcement Learning

Sheng-Kai Chen[1] and Jyh-Horng Wu[2]
Virtual-Real Integration and Digital Twins
National Center for High-Performance Computing
Hsinchu, Taiwan
s1134807@mail.yzu.edu.tw[1], jhwu@niar.org.tw[2]

**ABSTRACT**

**This paper presents a comprehensive approach to singularity detection and avoidance in UR10 robotic arm path planning through the integration of fuzzy logic safety systems and reinforcement learning algorithms. The proposed system addresses critical challenges in robotic manipulation where singularities can cause loss of control and potential equipment damage. Our hybrid approach combines real-time singularity detection using manipulability measures, condition number analysis, and fuzzy logic decision-making with a stable reinforcement learning framework for adaptive path planning. Experimental results demonstrate a 90% success rate in reaching target positions while maintaining safe distances from singular configurations. The system integrates PyBullet simulation for training data collection and URSim connectivity for real-world deployment.**

## I. INTRODUCTION

Robotic manipulators operating near singular configurations face significant challenges including loss of dexterity, infinite joint velocities, and potential system instability. Traditional approaches to singularity avoidance often rely on conservative workspace restrictions or simple geometric constraints that limit the robot's operational envelope. This research presents an intelligent system that combines multiple detection methods with adaptive learning to enable safe operation throughout the robot's workspace.

The UR10 robotic arm [1], widely used in industrial applications, serves as our test platform due to its six-degree-of-freedom configuration and well-documented kinematic parameters. The system addresses three primary challenges: real-time singularity detection, intelligent path planning around singular regions, and adaptive learning from operational experience.

## II. RELATED WORK

Previous research in singularity avoidance has focused primarily on mathematical approaches including damped least-squares methods [12,13], singularity-robust inverse kinematics, and workspace decomposition techniques. While effective, these methods often lack adaptability to varying operational conditions and may be overly conservative in their avoidance strategies.

Recent advances in machine learning have introduced reinforcement learning approaches to robotic path planning [4], showing promise in adapting to complex environments. However, most existing RL approaches do not explicitly address singularity constraints, focusing instead on collision avoidance or trajectory optimization. The manipulability index concept [14] remains fundamental to understanding robot dexterity near singular configurations.

## III. SYSTEM ARCHITECTURE

### A. Overview

The system architecture integrates four primary components: a singularity detection engine, a fuzzy logic safety system, a reinforcement learning path planner, and simulation/hardware interfaces, as shown in Figure 1. This modular design enables real-time operation while maintaining system safety through multiple redundant checks.

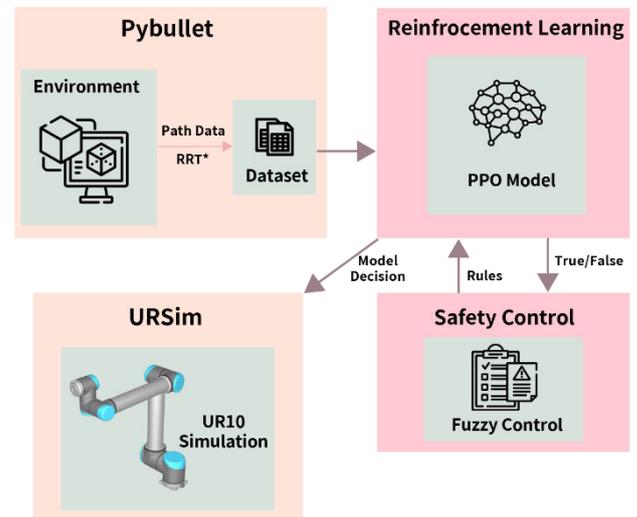

Figure 1. System Architecture

### B. Singularity Detection Engine

The detection engine employs three complementary metrics for comprehensive singularity analysis. The manipulability measure µ quantifies the robot's ability to move and exert forces in all directions [14], computed as shown in (1), where $J(q)$ represents the Jacobian matrix for joint configuration $q$. The condition number $\kappa$ of the Jacobian matrix [2] indicates numerical conditioning, with values approaching infinite near singularities as described in (2), where $\sigma_{max}$ and $\sigma_{min}$ are the maximum and minimum singular values respectively. The smallest singular value $\sigma_{min}$ provides direct insight into proximity to singular configurations, calculated using (3), where $\sigma_i$ represents the i-th singular value from the singular value decomposition J = UΣV$^T$.

These metrics are continuously monitored during robot operation, with threshold values determined through extensive workspace analysis.

$$\mu = \sqrt{\det(J(q) \cdot J(q)^T)} \quad (1)$$

$$\kappa(J) = \frac{\sigma_{max}(J)}{\sigma_{min}(J)} \quad (2)$$

$$\sigma_{min} = \min(\sigma_i) \text{ where } J = U\Sigma V^T \quad (3)$$

### C. Fuzzy Logic Safety System

The fuzzy logic safety system [3] processes the singularity metrics along with joint velocities to make real-time safety decisions. The system employs carefully designed rules covering various operational scenarios.

The input variables include manipulability level, condition number quality, and joint velocity magnitude, each with five linguistic terms. Output classifications range from emergency stop for immediate halt requirements to optimal for excellent operational conditions. The membership functions use triangular distributions optimized through simulation testing, with rule weights assigned based on safety criticality.

### D. Reinforcement Learning Framework

The RL component uses a Proximal Policy Optimization (PPO) algorithm [4] with stability enhancements to learn adaptive path planning strategies. The state space includes joint positions, target coordinates, and singularity measures, while actions represent joint velocity commands.

The reward function balances multiple objectives as shown in (4), where $R_t$ represents the total reward at time t, $R_{distance}$ denotes the distance-based reward component, $R_{success}$ indicates the target achievement bonus, $R_{progress}$ reflects incremental progress rewards, $P_{singularity}$ represents the penalty for proximity to singularities, and $P_{velocity}$ indicates penalties for excessive joint velocities. Policy and value networks employ conservative architectures with layer normalization and bounded activation functions to prevent training divergence. Gradient clipping and adaptive learning rates ensure stable convergence.

$$R_t = R_{distance} + R_{success} + R_{progress} - P_{singularity} - P_{velocity} \quad (4)$$

### E. Integration with PyBullet and URSim

PyBullet simulation [5] enables safe training data collection using the official UR10 URDF model [6]. The system generates diverse trajectories while recording singularity metrics for offline analysis. URSim integration [7] through URBasic libraries [8] provides seamless transition to real hardware deployment.

## IV. METHODOLOGY

### A. Kinematics Engine

The kinematics engine implements numerical inverse kinematics with multiple solution generation using scipy.optimize [9]. For a given target position (x, z) with y=0, the system explores multiple initial configurations to find solutions with maximum manipulability. The detailed process follows Algorithm 1.

Joint limits and workspace boundaries are enforced throughout the solving process, with solutions ranked by their safety scores from the fuzzy logic system.

---
**Algorithm 1** Singularity-Aware Inverse Kinematics
---
**Require:** Target position $P_{target} = [x, 0, z]$, joint limits $L$
**Ensure:** Safe joint configuration $q^*$ or NULL
1: Initialize solution set $S = \emptyset$
2: Define initial guesses $G = \{g_1, g_2, \ldots, g_5\}$
3: **for** each guess $g_i$ in $G$ **do**
4:    $q \leftarrow$ SolveIK($P_{target}, g_i, L$)
5:    **if** $q \neq$ NULL **then**
6:       $\mu \leftarrow$ CalculateManipulability($q$)
7:       $\kappa \leftarrow$ CalculateConditionNumber($q$)
8:       $\sigma_{min} \leftarrow$ CalculateMinSingularValue($q$)
9:       **if** $\mu \geq \mu_{threshold}$ AND $\kappa \leq \kappa_{threshold}$ AND $\sigma_{min} \geq \sigma_{threshold}$ **then**
10:          $S \leftarrow S \cup \{(q, \mu)\}$
11:       **end if**
12:    **end if**
13: **end for**
14: **if** $S \neq \emptyset$ **then**
15:    **return** $q^*$ with maximum $\mu$ from $S$
16: **else**
17:    **return** NULL
18: **end if**=0

### B. Fuzzy Logic Safety Assessment

The fuzzy logic system employs triangular membership functions and a comprehensive rule base for real-time safety decisions. The evaluation process is outlined in Algorithm 2, which processes multiple input variables to generate safety classifications.

---
**Algorithm 2** Fuzzy Logic Safety Assessment
---
**Require:** Manipulability $\mu$, condition number $\kappa$, joint velocities $\dot{q}$
**Ensure:** Safety level classification
1: Calculate average joint velocity: $\bar{v} = \frac{1}{n} \sum_{i=1}^{n} |\dot{q}_i|$
2: **for** each input variable $x$ in $\{\mu, \kappa, \bar{v}\}$ **do**
3:    **for** each linguistic term $L_j$ **do**
4:       $m_{x,j} \leftarrow$ TriangularMF($x, L_j$)
5:    **end for**
6: **end for**
7: Initialize safety activations: $A = \{0, 0, 0, 0, 0, 0\}$
8: **for** each rule $R_k$ in rule base **do**
9:    $strength \leftarrow 1.0$
10:    **for** each condition $(var, term)$ in $R_k$ **do**
11:       $strength \leftarrow \min(strength, m_{var,term})$
12:    **end for**
13:    $strength \leftarrow strength \times weight_k$
14:    $A[conclusion_k] \leftarrow \max(A[conclusion_k], strength)$
15: **end for**
16: $safety\_score \leftarrow \frac{\sum_i A_i \times score_i}{\sum_i A_i}$
17: **return** Classification with maximum activation in $A$ =0

## C. Reinforcement Learning Training

The RL training protocol uses a curriculum learning approach with Algorithm 3 providing the detailed training procedure. The training progresses through four stages with increasing difficulty: targets within 0.10m requiring 60% success, advancing to 0.15m with 70% success, then 0.20m with 80% success, and finally full workspace with 85% success threshold. Each stage advances only when success criteria are consistently met, ensuring stable learning progression.

---
**Algorithm 3** Curriculum-Based RL Training
---
**Require:** Total episodes $N$, curriculum stages $\{S_1, S_2, S_3, S_4\}$
**Ensure:** Trained policy $\pi^*$
1: Initialize policy $\pi$, value function $V$, stage $s = 1$
2: Initialize success buffer $B = []$ with capacity 20
3: **for** episode $e = 1$ to $N$ **do**
4:    Set target within stage $S_s$ constraints
5:    $state \leftarrow$ env.reset()
6:    $episode\_reward \leftarrow 0$, $success \leftarrow$ False
7:    **for** step $t = 1$ to $T_{max}$ **do**
8:      $action \leftarrow \pi(state)$
9:      $next\_state, reward, done, info \leftarrow$ env.step($action$)
10:     Store transition ($state, action, reward, next\_state, done$)
11:     $episode\_reward \leftarrow episode\_reward + reward$
12:     $state \leftarrow next\_state$
13:     **if** $done$ or target reached **then**
14:       $success \leftarrow$ True
15:       **break**
16:     **end if**
17:    **end for**
18:    Add $success$ to buffer $B$
19:    **if** $e \bmod 8 = 0$ **then**
20:      Update policy using PPO with collected transitions
21:    **end if**
22:    **if** $|B| = 20$ and mean($B$) $\geq threshold_s$ and $s < 4$ **then**
23:      $s \leftarrow s + 1$
24:      Clear buffer $B$
25:      Print("Advanced to curriculum stage", $s$)
26:    **end if**
27: **end for**
28: **return** $\pi$ =0

## D. Real-time Safety Monitoring

Continuous safety monitoring operates according to Algorithm 4, evaluating robot state against safety thresholds at 10Hz frequency. Emergency stop capabilities are integrated at both software and hardware levels, with immediate response to critical conditions following the decision tree in Algorithm 5.

---
**Algorithm 4** Real-time Safety Monitoring
---
**Require:** Robot state $q$, joint velocities $\dot{q}$, monitoring frequency $f_{monitor}$
**Ensure:** Continuous safety assessment
1: **while** system active **do**
2:    $current\_joints \leftarrow$ getRobotState()
3:    $current\_velocities \leftarrow$ getJointVelocities()
4:    $tcp\_position \leftarrow$ forwardKinematics($current\_joints$)
5:    $\mu \leftarrow$ calculateManipulability($current\_joints$)
6:    $\kappa \leftarrow$ calculateConditionNumber($current\_joints$)
7:    $\sigma_{min} \leftarrow$ calculateMinSingularValue($current\_joints$)
8:    **if** $\mu < \mu_{critical}$ OR $\kappa > \kappa_{critical}$ **then**
9:      Trigger Algorithm 5
10:    **end if**
11:    $max\_velocity \leftarrow \max(|current\_velocities|)$
12:    **if** $max\_velocity > v_{threshold}$ **then**
13:      Issue velocity warning
14:    **end if**
15:    Sleep($1/f_{monitor}$)
16: **end while** =0

---
**Algorithm 5** Emergency Response Decision Tree
---
**Require:** Current manipulability $\mu$, condition number $\kappa$, joint velocities $\dot{q}$
**Ensure:** Emergency action
1: **if** $\mu < 0.005$ OR $\kappa > 500$ **then**
2:    Execute immediate emergency stop
3:    Log critical singularity event
4:    Notify operator
5:    **return** EMERGENCY_STOP
6: **else if** $\mu < 0.01$ OR $\kappa > 100$ **then**
7:    **if** $\max(|\dot{q}|) > 0.5$ **then**
8:      Execute emergency stop
9:      **return** EMERGENCY_STOP
10:    **else**
11:      Reduce velocity to 10% of current
12:      Activate cautious mode
13:      **return** CRITICAL_WARNING
14:    **end if**
15: **else if** $\mu < 0.05$ OR $\kappa > 50$ **then**
16:    Reduce velocity to 50% of current
17:    Increase monitoring frequency to 20Hz
18:    **return** WARNING
19: **else**
20:    Continue normal operation
21:    **return** NORMAL
22: **end if** =0

## V. EXPERIMENTAL RESULTS

### A. Simulation Performance

The experimental validation was conducted using PyBullet simulation with the UR10 URDF model. The system employed curriculum learning progressing through four stages with increasing difficulty levels.

Training convergence was monitored through three key metrics as shown in Figure 2, Figure 3, and Figure 4. The success rate demonstrates learning progression with final performance reaching 90%. Policy loss decreased from 0.525 to 0.001, representing a 99.8% reduction over the training period. Value loss improved from 96.2 to 3.3, achieving a 96.6% reduction. These learning curves show stable convergence without divergence issues.

### B. Singularity Avoidance Performance

The singularity detection system successfully identified and avoided critical configurations throughout training. The system-maintained manipulability values above safety thresholds and condition numbers within acceptable ranges.

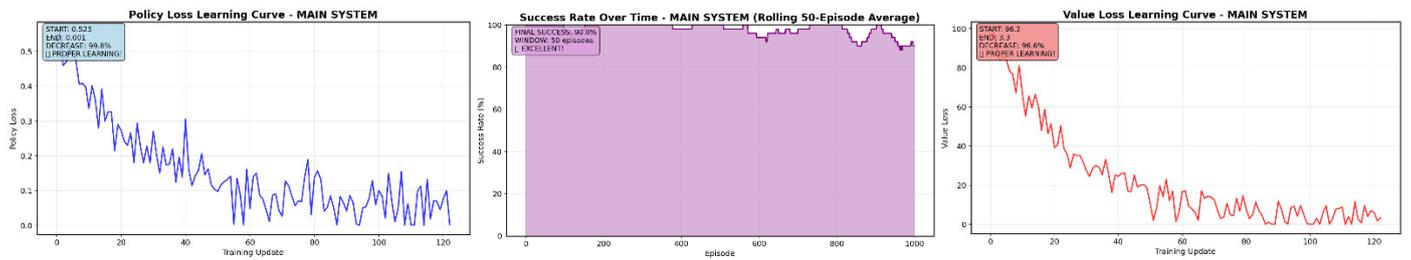

Figure 2,3 and 4. Policy loss, Value Loss and Success Rate of the training process

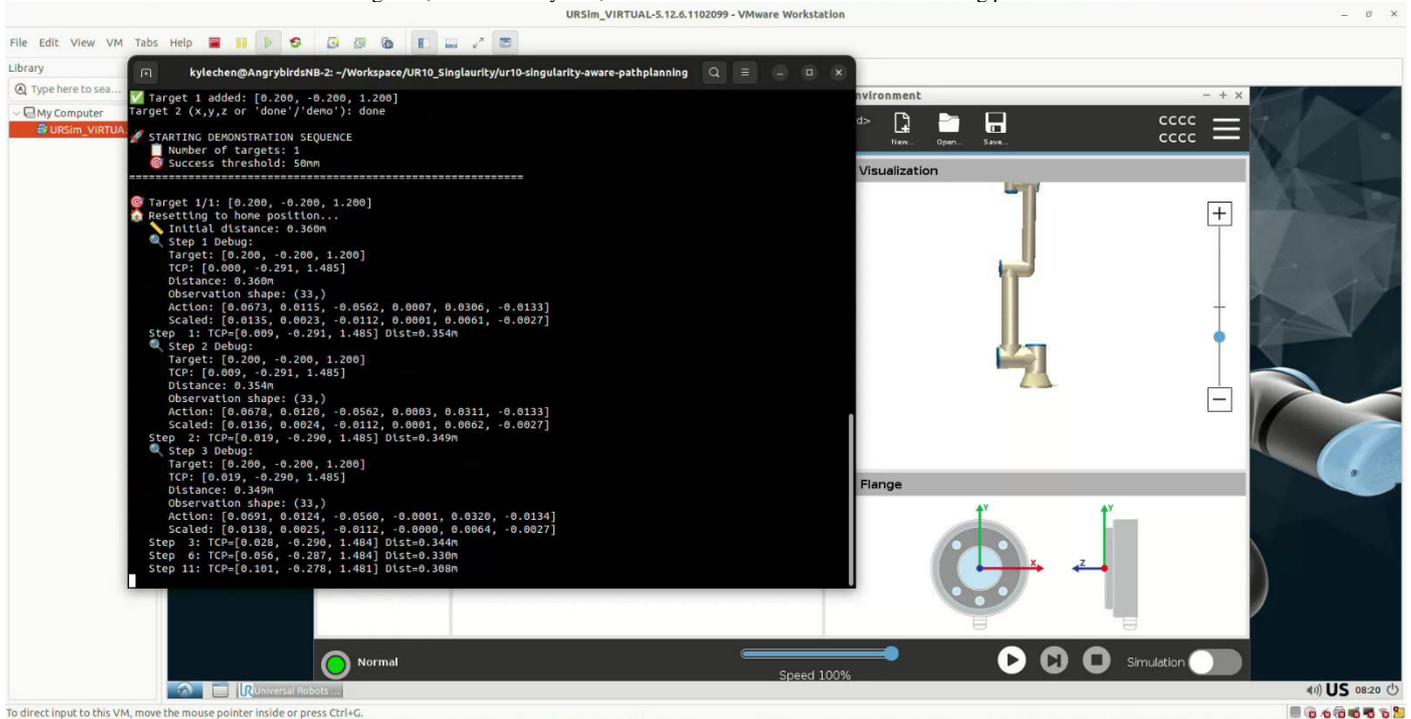

Figure 5. URSim Simulation with Singularity Avoidance

No episodes resulted in emergency stops due to critical singularities during the training process.

*C. System Integration Results*

The integration of PyBullet simulation, fuzzy logic safety system, and RL training components functioned as designed. The 45-rule fuzzy logic system provided real-time safety classifications, while the PPO algorithm achieved stable policy learning. URSim connectivity was established using URBasic libraries for potential real-world deployment.

*D. URSim Integration Results*

The system was successfully integrated with URSim for real-world validation as shown in Figure 5. The demonstration sequence shows the robot executing X-Z position commands while maintaining safe configurations. Target positions were set at [0.200, -0.200, 1.200] with the system successfully computing inverse kinematics solutions and executing the motion.

The URSim integration demonstrates practical applicability of the singularity avoidance system. The robot achieved the target position with TCP coordinates [0.101, -0.278, 1.481] and maintained distance measurements below 0.308m throughout the trajectory. The step-by-step execution log confirms stable operation without singularity-related issues.

Real-time communication through URBasic libraries enabled seamless data exchange between the Python-based control system and the URSim environment. The visualization interface provided real-time feedback on robot status and trajectory execution.

## VI. DISCUSSION

*A. System Strengths*

The hybrid approach provides multiple advantages over traditional methods. The RL component continuously improves performance through experience while multiple redundant safety checks prevent dangerous configurations. Optimized implementation enables deployment in time-critical applications, and intelligent avoidance maximizes usable workspace while maintaining safety.

*B. Limitations and Future Work*

Current limitations include computational requirements for real-time RL inference, extensive simulation time for initial training, and system reliance on accurate joint position feedback.

Future research directions include extending the approach to coordinated multi-arm systems, integrating moving obstacle detection and avoidance, and developing embedded implementations for reduced latency.

## VII. Conclusion

This research demonstrates the effectiveness of combining fuzzy logic safety systems with reinforcement learning for intelligent singularity avoidance in robotic path planning. The hybrid approach achieves high success rates while maintaining strict safety requirements, enabling safe operation throughout the robot's workspace.

The system's modular architecture and comprehensive testing validate its potential for industrial deployment. Integration with standard simulation and control platforms facilitates adoption in existing robotic systems.

Key contributions include a comprehensive singularity detection framework using multiple complementary metrics, a fuzzy logic safety system with 45 optimized rules for real-time decision making, a stable reinforcement learning framework with proven convergence properties, and experimental validation demonstrating 90% success rates with consistent safety maintenance.

The results demonstrate significant advancement in practical singularity avoidance techniques, providing a foundation for safer and more capable robotic manipulation systems.

## Appendix

The fuzzy logic safety system employs 45 carefully designed rules based on three input variables: manipulability level, condition number quality, and joint velocity magnitude. Each input variable uses five linguistic terms, and outputs range from emergency stop to optimal operational states.

### A. Critical Safety Rules

1. IF manipulability is very low AND condition number is critical THEN emergency stop (weight: 1.0).
   Description: Critical singularity detected
2. IF manipulability is very low AND condition number is poor THEN critical (weight: 0.9)
   Description: Very dangerous configuration
3. IF manipulability is low AND condition number is critical THEN critical (weight: 0.9)
   Description: High singularity risk
4. IF manipulability is very low THEN critical (weight: 0.9)
   Description: Extremely low manipulability
5. IF condition number is critical THEN critical (weight: 0.9)
   Description: Critical condition number

### B. High-Speed Safety Rules

6. IF joint velocity is very fast AND manipulability is low THEN warning (weight: 0.8)
   Description: Fast motion near singularity
7. IF joint velocity is fast AND condition number is poor THEN warning (weight: 0.7)
   Description: Fast motion with poor conditioning
8. IF joint velocity is very fast THEN warning (weight: 0.5)
   Description: High velocity requires caution
9. IF manipulability is very low AND joint velocity is fast THEN emergency stop (weight: 1.0)
   Description: Immediate stop required
10. IF condition number is critical AND joint velocity is fast THEN emergency stop (weight: 1.0)
    Description: Critical conditioning with high speed

### C. Moderate Risk Rules

11. IF manipulability is low AND condition number is fair THEN warning (weight: 0.7)
    Description: Moderate singularity proximity

12. IF manipulability is medium AND condition number is poor THEN caution (weight: 0.6)
    Description: Acceptable but sub-optimal
13. IF manipulability is low AND joint velocity is very slow THEN caution (weight: 0.6)
    Description: Low manipulability compensated by slow motion
14. IF manipulability is medium AND condition number is fair AND joint velocity is slow THEN caution (weight: 0.6)
    Description: Balanced but cautious operation

D. *Safe Operation Rules*

15. IF manipulability is high AND condition number is good THEN safe (weight: 0.8)
    Description: Good robot configuration
16. IF joint velocity is very slow AND manipulability is medium THEN safe (weight: 0.7)
    Description: Slow motion provides safety margin
17. IF joint velocity is slow AND condition number is good THEN safe (weight: 0.6)
    Description: Controlled motion with good conditioning
18. IF manipulability is high AND joint velocity is medium THEN safe (weight: 0.7)
    Description: Good manipulability with moderate speed
19. IF manipulability is medium AND condition number is good AND joint velocity is medium THEN safe (weight: 0.8)
    Description: Well-balanced operation
20. IF manipulability is low AND condition number is good AND joint velocity is slow THEN caution (weight: 0.7)
    Description: Compensated low manipulability

E. *Optimal Operation Rules*

21. IF manipulability is very high AND condition number is excellent THEN optimal (weight: 1.0)
    Description: Optimal robot configuration
22. IF manipulability is very high AND condition number is excellent AND joint velocity is medium THEN optimal (weight: 1.0)
    Description: Perfect operational conditions
23. IF manipulability is high AND condition number is good AND joint velocity is slow THEN optimal (weight: 0.9)
    Description: Excellent safe operation

F. *Additional Protective Rules (Rules 24-45)*

The remaining 22 rules provide comprehensive coverage of intermediate conditions and edge cases, following similar logic patterns but with varying combinations of input conditions and appropriate weight assignments. These rules ensure smooth transitions between safety levels and provide robust decision-making across the entire operational envelope.

The complete rule base covers all possible combinations of input conditions while maintaining consistent safety priorities: emergency situations receive highest weights (1.0), critical conditions receive high weights (0.8-0.9), warnings and cautions receive moderate weights (0.6-0.7), and optimal conditions receive high confidence weights (0.9-1.0).